\documentclass[10pt,twocolumn,letterpaper]{article}

\usepackage{cvpr}
\newcommand{\keypoint}[1]{\vspace{0.01cm}\noindent\textbf{#1}}

\usepackage{pifont}
\newcommand{\cmark}{\ding{51}}
\newcommand{\xmark}{\ding{55}}

\usepackage{multirow}
\usepackage[dvipsnames]{xcolor, colortbl}

\newcommand{\modelname}{3DrawAgent}

\definecolor{cvprblue}{rgb}{0.21,0.49,0.74}
\usepackage[pagebackref,breaklinks,colorlinks,allcolors=cvprblue]{hyperref}

\def\paperID{18396}
\def\confName{CVPR}
\def\confYear{2026}

\title{\modelname: Teaching LLM to Draw in 3D with Early Contrastive Experience}

\author{
    Hongcan Xiao$^{1}$ \quad Xinyue Xiao$^{1,2}$ \quad Yilin Wang$^{1}$ \quad Yue Zhang$^{3}$ \quad Yonggang Qi$^{1\dagger}$ \\
    $^1$Beijing University of Posts and Telecommunications \quad $^2$Jiangnan University \quad $^3$HaoHan Data \\
    \begin{tabular}{ccc}
        {\small \tt \{xiaohc, wangyilin2022\}@bupt.edu.cn} & {\small \tt 1193220402@stu.jiangnan.edu.cn} & {\small \tt zhangyue@haohandata.com}
    \end{tabular}
}

\begin{document}
\maketitle
\insert\footins{\noindent\footnotesize $^{\dagger}$ Corresponding author: qiyg@bupt.edu.cn}
\begin{abstract}
Sketching in 3D space enables expressive reasoning about shape, structure, and spatial relationships, yet generating 3D sketches through natural language remains a major challenge. In this work, we introduce 3DrawAgent, a training-free, language-driven framework for 3D sketch generation that leverages large language models (LLMs) to sequentially draw 3D Bezier curves under geometric feedback. Unlike prior 2D sketch agents, our method introduces a relative experience optimization strategy that adapts the recently proposed Group Reward Policy Optimization (GRPO) paradigm. Instead of relying on explicit ground-truth supervision, we construct pairwise comparisons among generated sketches, with each pair consisting of a relatively better and a worse result based on CLIP-based perceptual rewards and LLM-based fine-grained qualitative assessment. These experiences are then used to iteratively refine the prior knowledge of 3D drawing, enabling black-box reinforcement of the model's 3D awareness. This design allows our model to self-improve its spatial understanding and drawing quality without parameter updates. Experiments show that 3DrawAgent can generate complex and coherent 3D Bezier sketches from diverse textual prompts, exhibit emergent geometric reasoning, and generalize to novel shapes, establishing a new paradigm for advancing the field of training-free 3D sketch intelligence.
\end{abstract}

\section{Introduction}
\label{sec:intro}

Sketching has long served as a universal medium for conceptualization and communication. From early drafts to modern graphics, it allows humans to externalize complex spatial reasoning through a few expressive strokes.

Recent progress in large language models (LLMs) and multimodal systems has dramatically expanded the landscape of content creation and human-AI interaction. Yet, capturing the 3D and structurally consistent nature of human sketching remains a major challenge. While language-driven sketch generation has shown promise in 2D contexts~\cite{frans2022clipdraw, jain2022vectorfusion}, generating 3D sketches that reflect spatial relationships and geometric intent is still largely unexplored.

\begin{figure}
  \centering
  \includegraphics[width=\linewidth]{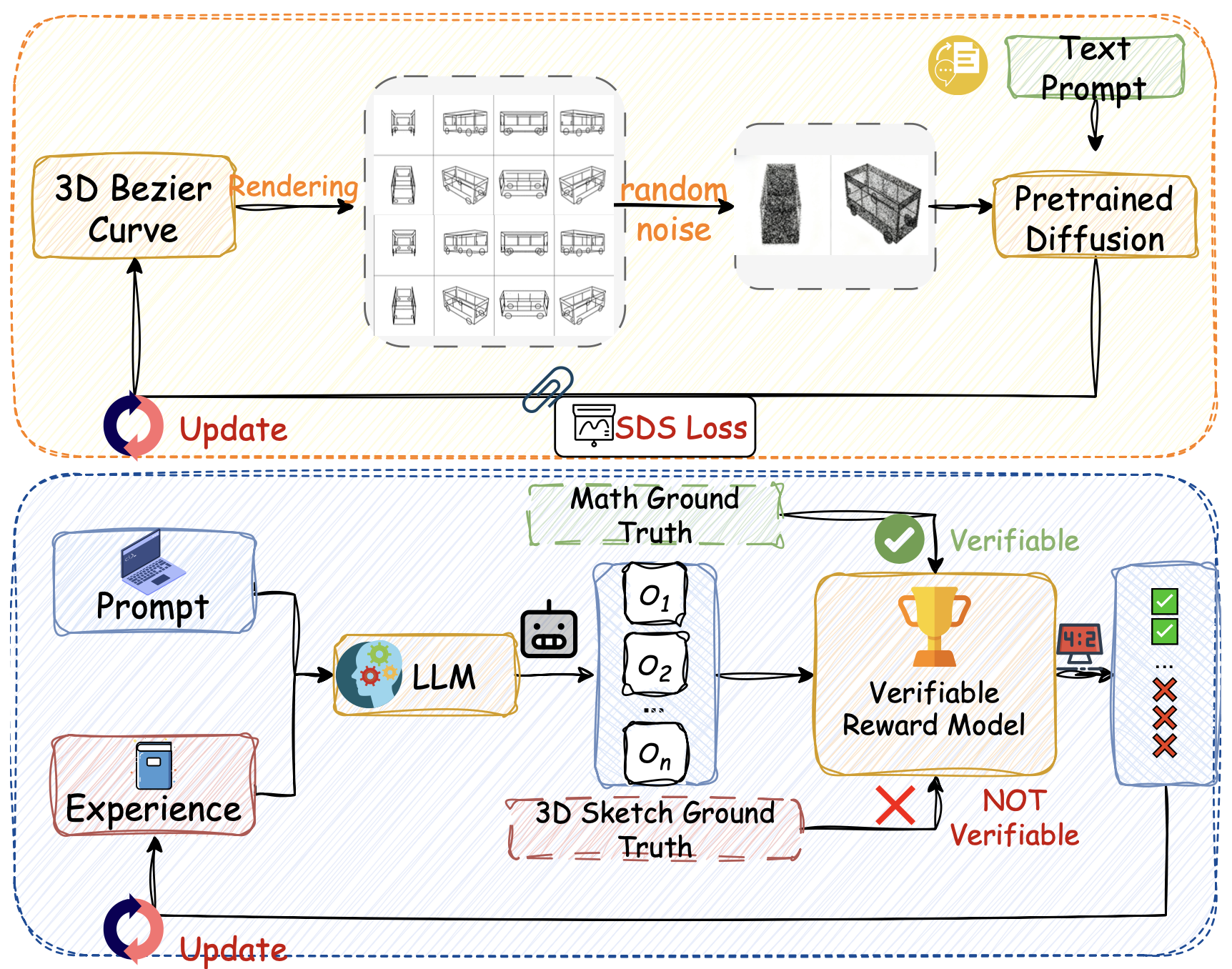}
  \caption{\textbf{Top}: Prior works typically rely on pre-trained diffusion models as 3D priors. \textbf{Bottom:} Our work performs training-free 3D sketch generation by refining an LLM's spatial reasoning.}
  \label{fig:open}
\end{figure}

Existing approaches to 3D shape generation, such as diffusion-based~\cite{zeng2022lion, poole2022dreamfusion, lin2023magic3d} or neural implicit methods~\cite{Park_2019_CVPR_deepsdf}, require explicit geometry supervision or extensive retraining. These paradigms, while powerful, are computationally intensive and lack interactivity. Meanwhile, recent progress in language-driven sketching, exemplified by SketchAgent~\cite{vinker2025sketchagent}, demonstrates that off-the-shelf multimodal LLMs can produce sequential vector drawings purely through in-context prompting. However, such methods remain confined to the image plane: they operate in 2D coordinate space and cannot reason about depth, projection, or geometric consistency. Moreover, while training-free optimization techniques such as training-free GRPO~\cite{training_free_grpo} enable lightweight model adaptation, they typically rely on scalar rewards or ground-truth references, both of which are impractical for open-ended creative tasks like sketching.

In this work, we introduce \modelname, a training-free, language-driven framework for 3D sketch generation. Our method leverages the sequential reasoning capability of LLMs~\cite{yao2022react} to draw 3D Bezier curves step by step, effectively extending the notion of language-driven sketching into 3D space. To equip the model with spatial awareness without parameter updates, we propose a contrastive experience optimization strategy inspired by training-free GRPO~\cite{training_free_grpo}. Instead of relying on ground-truth 3D sketches, our framework constructs pairwise experiences among generated results, identifying relatively better and worse sketches through a combination of CLIP-based perceptual evaluation~\cite{radford2021clip} and LLM-based fine-grained qualitative judgment~\cite{zheng2023judging}. These experiences are used to iteratively refine the in-context prompts, thus realizing a form of black-box reinforcement prompt tuning that strengthens the model's understanding of 3D geometry.

This design introduces a new training-free adaptation paradigm: the LLM not only generates sketches but also learns from its own outputs via self-assessment. Through iterative feedback, our model progressively improves its drawing quality and spatial reasoning ability, capturing features such as depth coherence, symmetry, and curvature alignment. Experimental results show that our method can generate coherent 3D Bezier sketches from diverse textual prompts, generalize to unseen shapes, and even exhibit emergent 3D reasoning capabilities.

Our main contributions are as follows: (i) A language-driven 3D sketching framework that enables LLMs to generate Bezier-based 3D sketches sequentially and interactively. (ii) A relative experience optimization mechanism that extends training-free GRPO to pairwise, self-supervised prompt reinforcement without ground-truth supervision. (iii) A hybrid reward design combining CLIP-based perceptual feedback and LLM-based qualitative assessment, allowing fine-grained evaluation of spatial and structural quality.

\section{Related Work}
\keypoint{Language-Driven Drawing Agents.} Language-guided sketching aims to bridge symbolic reasoning and visual abstraction. Early systems such as SketchRNN~\cite{ha2018a} and CLIPDraw~\cite{frans2022clipdraw} explored language-conditioned 2D sketch generation through sequence modeling or CLIP-based optimization. More recently, SketchAgent~\cite{vinker2025sketchagent} demonstrated that off-the-shelf multimodal large language models (LLMs) can serve as drawing agents, producing sequential vector sketches~\cite{carlier2020deepsvg} purely from in-context examples and dialogue. These advances mark a step toward natural, conversational drawing systems. However, existing methods are confined to 2D canvases, i.e., they operate in planar coordinate spaces and lack awareness of depth, geometry, and structure. As a result, generated sketches often fail to capture the spatial consistency that human sketches naturally convey. In contrast, our model extends this paradigm into three-dimensional space. By introducing a Bezier-based 3D sketch language, we enable LLMs to reason about geometry and structure while drawing step by step. This moves beyond flat sketching toward language-driven 3D structural reasoning.

\keypoint{Training-Free Foundation Model Adaptation.} Recent research has sought ways to adapt foundation models without parameter updates, relying on iterative feedback rather than explicit training. Methods such as ~\cite{madaan2023self, shinn2023reflexion} allow language models to improve through textual self-assessment, while training-free GRPO~\cite{training_free_grpo} generalizes reinforcement learning (RL~\cite{ouyang2022training, deepseek-math}) to training-free, black-box optimization. Despite their efficiency, these techniques generally depend on scalar rewards or ground-truth references, which are insufficient for open-ended creative domains such as drawing or design. We address this limitation by introducing a relative experience optimization strategy. Instead of defining absolute correctness, we form pairwise experiences between generated sketches, thereby identifying which result better captures 3D geometry. A hybrid feedback system combines CLIP-based perceptual scores~\cite{radford2021clip} (quantifying shape-text alignment) with LLM-based qualitative evaluation~\cite{zheng2023judging} (capturing fine-grained compositional quality). This design transforms GRPO into a black-box reinforcement prompt tuning process, enabling self-improvement of 3D sketching ability without any gradient-based update or external supervision.

\keypoint{3D Sketch Representation and Modeling.} 3D sketch modeling lies at the intersection of geometric reasoning and conceptual design, aiming to represent shapes through curves, strokes, or wireframes that encode both topology and spatial layout. Prior works such as DeepCAD~\cite{wu2021deepcad} and SketchGraphs~\cite{SketchGraphs} represent 3D CAD structures via parametric curve sequences, while subsequent works, such as Sketch2CAD~\cite{li2020sketch2cad} and Text2cad~\cite{khan2024textcad}, integrate language or image guidance to produce structured sketches. Other methods, e.g., \cite{zhou2019learning} and \cite{liu2021pcwf}, reconstruct 3D edge structures from multi-view or point cloud cues. In addition, 3D Gaussian Splatting (3DGS) has also been adopted as a new paradigm for curve and edge reconstruction, with methods such as SketchSplat~\cite{Ying_2025_ICCV}, EdgeGaussians~\cite{Chelani_2025_EdgeGaussians} and CurveGaussian~\cite{gao2025curveawaregaussiansplatting3d}, showing how sketch-like primitives can be faithfully embedded into Gaussian-based neural rendering frameworks.
More recently, Bezier and parametric-curve-driven 3D generative modeling has gained traction, as seen in 3Doodle~\cite{10.1145/3658156}, Diff3DS~\cite{zhang2025diffds}, ViewCraft3D~\cite{wang2025viewcraftd}, and Dream3DVG~\cite{Li_2025_CVPR}. These methods jointly optimize differentiable curve primitives and multi-view consistency, advancing editable and semantically controlled shape synthesis.
Although these techniques excel at modeling explicit geometry, they typically rely on computationally intensive per-instance optimization (e.g., Score Distillation Sampling~\cite{poole2022dreamfusion}) and lack sequential planning capabilities, limiting their capacity for fast, interactive 3D sketch generation. In contrast, our approach leverages LLM-driven planning~\cite{yao2022react} and CLIP-based self-evaluation to perform language-conditioned 3D sketch generation without predefined motion priors or ground-truth supervision.

\begin{figure*}[t]
  \centering
  \includegraphics[width=\linewidth]{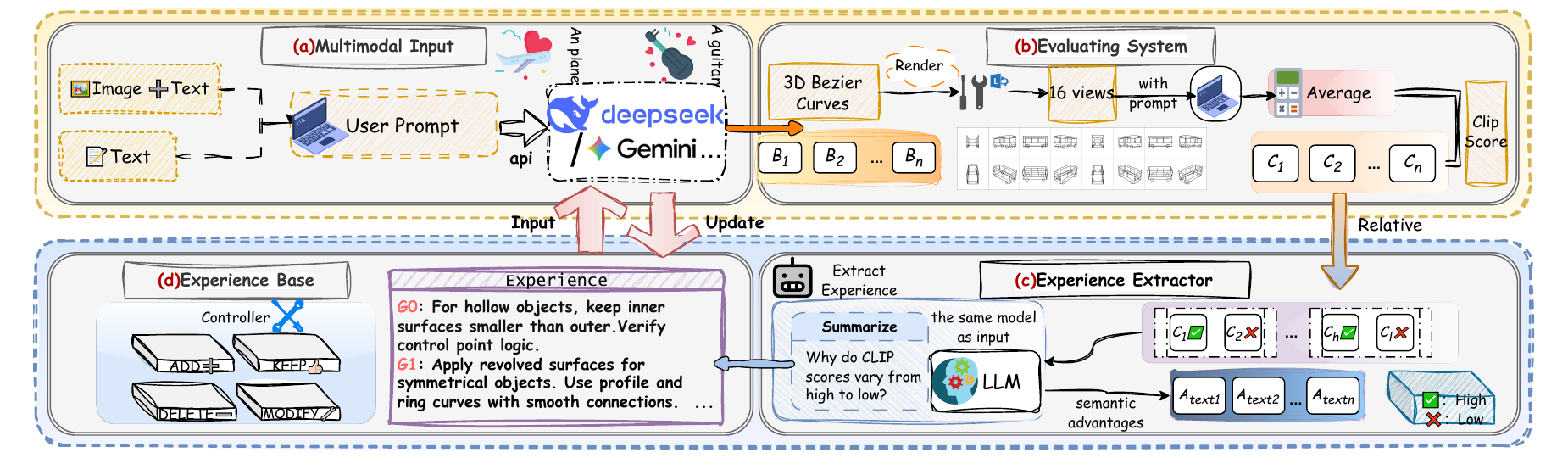}
  \caption{Framework Overview. Given a text prompt, our framework uses an LLM to autoregressively generate 3D Bezier curves. Each generated sketch is evaluated with a CLIP-based model to produce quality scores, forming contrastive pairs that teach the LLM which sketches are better or worse. These insights are accumulated into an experience library, which is then leveraged to guide subsequent language-driven 3D sketch generation, enabling coherent, semantically aligned, and spatially consistent 3D drawings.}
  \vspace{-0.5cm}
  \label{fig:overview}
\end{figure*}

\section{Method}

\subsection{Problem Setup}
We aim to enable language-driven 3D sketch generation without any 3D supervision or model fine-tuning. Specifically, given a textual description $\mathcal{T}$ and an optional reference image, our model generates a 3D sketch represented by a set of 3D Bezier curves $\mathcal{S}$:
\begin{align}
    \mathcal{S} &= \{ \mathbf{C}_1, \mathbf{C}_2, \ldots, \mathbf{C}_N \}, \\
    \mathbf{C}_i &= \text{Bezier}(\mathbf{P}_i^{(0)}, \mathbf{P}_i^{(1)}, \mathbf{P}_i^{(2)}, \mathbf{P}_i^{(3)})
    \label{eq:bezier}
\end{align}
where each curve $\mathbf{C}_i$ is parameterized by four 3D control points $\mathbf{P}_i^{(k)}\in \mathbb{R}^3$. There are mainly three stages to achieve the goal: (i) Language-conditioned 3D curve generation via in-context prompting of an LLM; (ii) Contrastive experience learning from a training-free GRPO-inspired experience accumulation process; and (iii) Language-driven 3D sketch generation using the obtained experience. An overview of our framework is shown in Figure~\ref{fig:overview}; we detail each key module in the following.

\subsection{Language-driven 3D Sketch Planning}

\keypoint{3D Sketch Representation using Language.} We first define a 3D sketch language that expresses drawing actions in a symbolic form understandable by LLMs. Similar to SketchAgent \cite{vinker2025sketchagent}, each action takes the form:
\begin{equation}
     a_t = \text{draw\_bezier}\left[\left(\mathbf{P}^{(0)}, \mathbf{P}^{(1)}, \mathbf{P}^{(2)}, \mathbf{P}^{(3)}\right)\right]
    \label{eq:action}
\end{equation}
and the full sketch is a sequence $\mathcal{A}=\{a_1,a_2,\dots,a_N\}$, where $N$ is the number of 3D strokes. Through in-context examples $c=(\mathcal{T}_i,\mathcal{A}_i)$, the model learns to map from text to 3D actions:
\begin{equation}
    p_\theta(\mathcal{A}|\mathcal{T},c)
\end{equation}
where $\theta$ refers to frozen foundation LLMs (e.g., \texttt{Gemini}~\cite{comanici2025gemini}, \texttt{DeepSeek}~\cite{deepseekai2024deepseekv32}, etc.). Unlike SketchAgent, our action space includes depth and spatial continuity constraints, allowing the model to reason about 3D layout.

\keypoint{Prompting Design.}
An effective in-context prompt is critical. It teaches a frozen LLM how to ``think like a 3D artist'' and produce LLM-parseable Bezier outputs. Concretely, our prompt design has the following components: (a) \emph{Role Instruction:} A short role statement (e.g., ``You are a professional 3D artist...'') primes the model to adopt the desired style and level of precision. This biases generation toward geometry-aware, procedural outputs rather than free-form prose. (b) \emph{Output Format Specification:} This is to remove ambiguity and make downstream parsing deterministic with explicit, unambiguous formatting rules (e.g., ``only a Python list within \texttt{<curves>}...\texttt{</curves>}''). (c) \emph{Data Type Constraints:} These provide exact type/shape constraints (e.g., each curve = list of 4 control points, each control point = list of 3 floats) to enforce that the LLM expresses geometry in a fixed symbolic representation compatible with the renderer and verifier. (d) \emph{Coordinate System:} This is to specify the canvas where LLM should draw (i.e., location, scale, and orientation), ensuring consistency across candidates and enabling meaningful multi-candidate comparisons without ambiguity. (e) \emph{Ground Truth Example:} Similar to SketchAgent, a small set of explicitly correct examples mapping \texttt{prompt} $\to$ \texttt{<curves>} is the primary mechanism for in-context learning~\cite{brown2020language_icl} of both semantic mapping and strict formatting. (f) \emph{Edge-case Rules:} Explicit bans (no comments, no variable assignments, no extra text inside the delimiter) reduce failure modes that corrupt parsers. Please refer to the supplementary for more details.

\keypoint{LLM as Spatial Planner.}
Given a novel text description $\mathcal{T}$, the frozen large language model (LLM) acts as a spatial planner~\cite{yao2022react} that generates an initial sequence of 3D drawing actions in a single forward process. Benefiting from its native language reasoning capability, the LLM interprets textual semantics (e.g., object parts, topology, and relative layout) and translates them into 3D Bezier curve primitives. Specifically, conditioned on the in-context examples $c=(\mathcal{T}_i,\mathcal{A}_i)$, the model predicts
\begin{equation}
    \hat{\mathcal{A}} = \arg\max_{\mathcal{A}} p_\theta(\mathcal{A}|\mathcal{T}, c),
\end{equation}
where $\hat{\mathcal{A}}$ denotes the generated 3D action sequence and $\theta$ are the frozen LLM parameters. Unlike prior models trained with explicit 3D supervision, our LLM-based planner leverages prompt engineering and accumulated experience to self-refine its spatial reasoning.

\keypoint{3D Parsing and Rendering.}
Each LLM output is a structured text describing a set of 3D Bezier curves. We design a lightweight \emph{parser-renderer pipeline} that converts this symbolic representation into renderable 3D sketches. Specifically, our parser extracts the content within the predefined delimiters (e.g., \texttt{<curves>}... \texttt{</curves>}), validates syntax and numerical ranges, and transforms each curve specification
$\mathcal{C}_i = \{P_0, P_1, P_2, P_3\}, P_j \in \mathbb{R}^3$ into parametric form:
\begin{equation}
    B_i(t) = \sum_{j=0}^{3} \binom{3}{j} (1-t)^{3-j} t^j P_j,\quad t\in[0,1]
\end{equation}
Afterwards, all curves are passed to a differentiable renderer~\cite{li2020differentiable} that supports orthographic and perspective projection. We employ depth-aware curve rasterization,
which yields consistent 2D views for CLIP-based scoring and multi-view visualization, while preserving the continuous geometry needed for gradient-free optimization. This parser-renderer bridge enables a seamless loop between symbolic reasoning (in language space) and geometric validation (in visual space), ensuring that every textual output can be objectively assessed and refined.

\subsection{Contrastive Knowledge Extraction}

\keypoint{Training-free GRPO vs Our Adaptation.}
We build on \emph{training-free GRPO}~\cite{training_free_grpo} to equip our model with 3D spatial reasoning capability without any parameter updates. Essentially, training-free GRPO optimizes prompt structures via group-based relative evaluation, i.e., it requires a group of candidate generations, from which the model identifies a clear winner with the reward model and computes relative semantic advantages against other losers within the same group. This process captures fine-grained comparative feedback but depends on {coherent group statistics and sufficient sample diversity}.

\emph{Our Adaptation.}
Instead of enforcing group-wise comparison, we generalize the paradigm to a \emph{pairwise contrastive experience} setting.
{We assume that relative quality signals can be extracted even from randomly paired generations $(o_i, o_j)$ as long as their perceptual difference is non-trivial. }
Intuitively, by iteratively integrating such pairwise experiences into the in-context prompt, the model gradually refines its internal geometric reasoning and drawing behavior. Moreover, unlike the original GRPO~\cite{deepseek-math}, which estimates a numerical group advantage, our contrastive formulation only relies on \emph{relative comparisons}, requiring no ground-truth sketches, gradient updates, or structured group rollouts.
This design turns the open-ended text-to-3D sketch generation task into a flexible form of black-box reinforcement prompt tuning, where our model learns to improve purely from self-produced contrastive feedback.

\keypoint{CLIP-based Scoring.}
To obtain perceptual signals for 3D sketches without ground-truth geometry, we employ a pre-trained CLIP~\cite{radford2021clip} model to score the rendered sketches. Each 3D sketch $\mathcal{S}$ is projected into multiple 2D views $\{I_v\}$ via our differentiable renderer, and the CLIP similarity between each $I_V$ and the textual description $\mathcal{T}$ is computed:
\begin{equation}
r_{\text{CLIP}} = \frac{1}{V} \sum_{v=1}^{V} \text{cos}\left(\text{E}_{\text{I}}(I_V), \text{E}_{\text{T}}(\mathcal{T})\right),
\end{equation}
where $\text{E}_{\text{I}}$ and $\text{E}_{\text{T}}$ are the CLIP image-encoders and text-encoders, respectively. This produces a perceptual alignment score that reflects how well the generated 3D structure visually corresponds to the text.

\keypoint{Contrastive Pairs.}
Given a batch of generated sketches $\{\mathcal{S}_i\}$ with their respective CLIP scores $\{r_i\}$, we construct pairwise contrastive experiences $(\mathcal{S}_i^+, \mathcal{S}_j^-)$ such that $r_i > r_j$. Each pair encodes a relative preference signal rather than an absolute label, making the approach inherently supervision-free. This flexible construction supports self-comparison across generations and temporal accumulation of experiences from different prompts, effectively serving as a non-parametric reward model.

\keypoint{LLM as Semantic Advantage Judge.}
In our case, the LLM serves as a \emph{semantic advantage estimator}~\cite{zheng2023judging}, analogous to the role of textual advantage extraction in training-free GRPO~\cite{training_free_grpo}. Given a contrastive pair of generated 3D sketches $(\mathcal{S}_i^+, \mathcal{S}_j^-)$, we prompt the LLM to perform comparative reasoning as:
\begin{equation}
    A^{\text{text}} = \texttt{LLM}(p_{\text{judge}}, \mathcal{T}, \mathcal{S}_i, \mathcal{S}_j, \mathcal{E}),
\end{equation}
where $p_{\text{judge}}$ is a reasoning template asking the model to articulate \emph{why} one sketch is better or worse in terms of structural integrity, spatial continuity, or geometric plausibility, given the current experiential knowledge $\mathcal{E}$.

\keypoint{3D Drawing Knowledge Extraction.}
The obtained $A^{\text{text}}$ thus functions as a natural-language \emph{semantic advantage}, encapsulating the reasoning patterns that lead to higher perceptual quality.
Following training-free GRPO, this advantage is then used to refine the 3D Drawing Knowledge, i.e., experience library $\mathcal{E}$, through discrete editing operations:
\begin{align}
    \mathcal{E} &\leftarrow \texttt{Update}(\mathcal{E}, A^{\text{text}}), \\
    \texttt{Update} &\in \{\text{Add}, \text{Delete}, \text{Modify}, \text{Keep}\}.
\end{align}
Through continuous accumulation of such comparative experiences, the LLM gradually internalizes 3D-aware drawing strategies, e.g., maintaining consistent topology, improving curve continuity, and ensuring spatial symmetry, without any gradient-based updates. This mechanism enables the model to evolve its geometric reasoning purely from self-reflective feedback, transforming each contrastive pair into actionable 3D sketch knowledge.

\subsection{3D Drawing with Extracted Experience}
Once the external experience library $\mathcal{E}$ that encodes transferable knowledge of geometric plausibility and spatial continuity is obtained, we formulate the drawing process given any novel text prompt $\mathcal{T}$ as conditional generation:
\begin{equation}
    o = p_\theta(o \mid \mathcal{T}, \mathcal{E}),
\end{equation}
where $p_\theta$ denotes the frozen LLM conditioned on both the text query $\mathcal{T}$ and the accumulated experience $\mathcal{E}$.
$\mathcal{E}$ is injected into the model's context window as an additional prompt segment that summarizes key spatial principles, e.g., maintain consistent curvature continuity across Bezier segments and preserve closed topology for symmetric objects.
The LLM then autoregressively produces a complete 3D sketch description in one pass, following the strict format defined in the prompt. It outputs a single Python list wrapped in \texttt{<curves>} and \texttt{</curves>}, encoding all Bezier control points, which are extracted, validated, and converted by our parser into numerical points for rendering.

\section{Experiments}
\subsection{Experimental Setup}
\keypoint{Implementation Details.} For 3D sketch rendering, we use a custom batched renderer built on top of the \texttt{pydiffvg} differentiable rendering library~\cite{li2020differentiable}. During evaluation, the renderer loads 16 fixed camera viewpoints and projects all 3D Bezier curves onto a $512 \times 512$ canvas using perspective projection.
Our framework interacts with two types of Large Language Models (LLMs): the open-source \texttt{DeepSeek-V3.2-Exp~\cite{deepseekai2024deepseekv32}} and the commercial \texttt{Gemini-2.5Pro~\cite{comanici2025gemini}}. During contrastive experience extraction, we sample $K=5$ candidate sketches per query to form the contrastive group, using a temperature of 0.7 to encourage output diversity. For the final inference stage, we adopt a lower temperature of 0.3 and report Pass@1 (i.e., the success rate of the first generation attempt) performance to emphasize high-quality generations.
Our training-free method relies on LLM APIs, requiring minimal compute; all experiments run on a single RTX 3090 GPU.

\begin{table*}[t]
\begin{center}
\caption{Comparison results on Text-to-3D (category- and fine-grained) and Image-to-3D generation. ``-'': not reported. }
\label{table:main_results}
\setlength{\tabcolsep}{1.3pt}
\scalebox{1}{
\begin{tabular}{lccccccc}
\toprule
\multirow{2}{*}{\bf Method} & \multirow{2}{*}{Train}   & \multicolumn{2}{c}{\bf Text-to-3D (Category)} & \multicolumn{2}{c}{\bf Text-to-3D (Fine-Grained)} & \multicolumn{2}{c}{\bf Image-to-3D} \\
\cline{3-8}
& &  CLIP-S$_T$ & AES  &  CLIP-S$_T$ & AES & CLIP-S$_I$ & AES \\
\midrule
{Diff3DS \cite{zhang2025diffds}} & \cmark  & 0.648 & 3.791  & 0.650 & 3.770  & 0.865 & 3.828 \\
{3Doodle \cite{10.1145/3658156}} & \cmark  & - & - & - & -  &  0.869 &  4.264 \\
{Dream3DVG \cite{Li_2025_CVPR}} & \cmark  & 0.660 & 4.150 & 0.670 & 4.174  & - & - \\

\midrule
3DrawAgent (\texttt{DeepSeek-V3.2})  & \xmark  & 0.643 & 4.108 & 0.664 & 4.146  & - & - \\
3DrawAgent (\texttt{Gemini-2.5 Pro})  & \xmark  & 0.649 & 4.161 & 0.669 & 4.175  & 0.873 & 4.255 \\

\bottomrule
\end{tabular}
}
\vspace{-1cm}
\end{center}
\end{table*}

\begin{figure*}
    \centering
    \includegraphics[width=\linewidth]{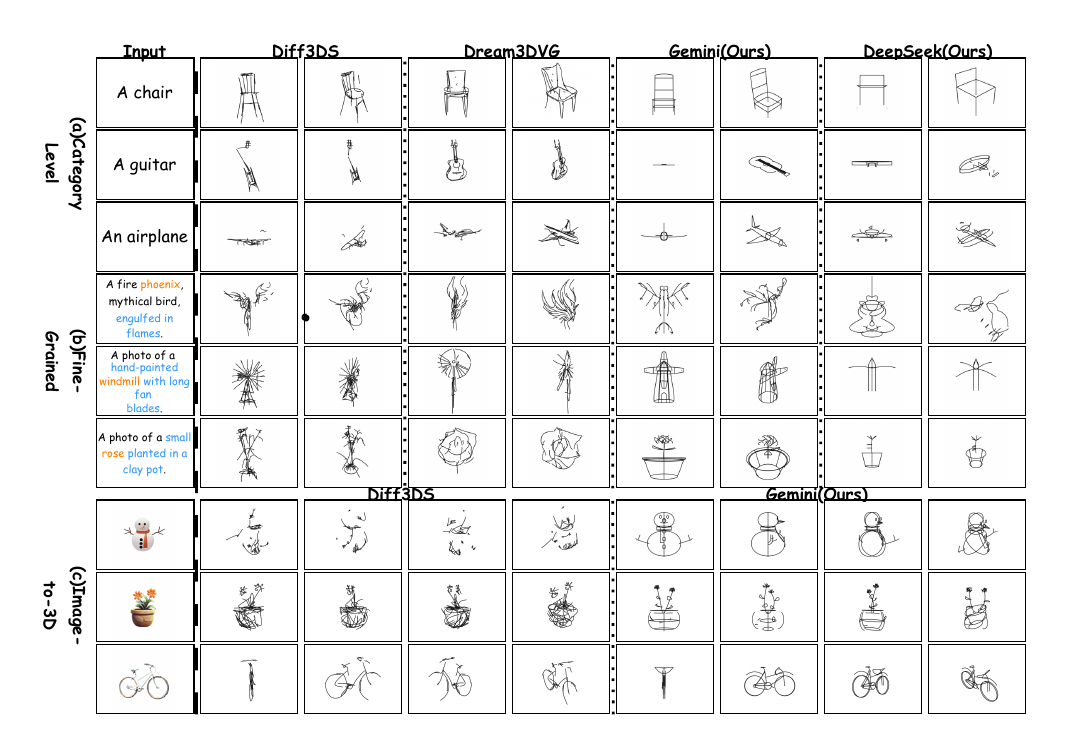}
    \vspace{-1cm}
    \caption{Comparison results on \textbf{(a) Category-level}, \textbf{(b) Fine-grained text-to-3D generation}, \textbf{(c) Image-to-3D generation}.}
    \vspace{-0.5cm}
    \label{fig:qualitative}
\end{figure*}

\keypoint{User Prompts.} To systematically evaluate our 3D sketch generation capability, we construct a new benchmark consisting of diverse user prompts derived from the object categories of ModelNet40~\cite{Zhirong15CVPR_3d_shapenets} and QuickDraw~\cite{jongejan2016quickdraw}. We select representative categories from both datasets that span a wide spectrum of 3D sketchable objects, ranging from rigid CAD-like geometries (e.g., chairs, tables, sofas, monitors, beds, cars, airplanes, lamps, bookshelves) to free-form, everyday hand-drawn concepts (e.g., cats, dogs, bicycles, trees, cups, houses, boats).
In addition to the above prompts, we further incorporate textual and visual prompts from Diff3DS~\cite{zhang2025diffds} to test the robustness and generality of our method under more complex input conditions: \emph{Diff3DS-Text} contains a collection of 28 complex and highly descriptive textual prompts, often involving fine-grained object properties or abstract conceptual descriptions. \emph{Diff3DS-Image} provides 37 reference images to evaluate image-to-3D performance.

\keypoint{Competitors.} We compare our method against three state-of-the-art 3D sketch generation methods: \textbf{Diff3DS}~\cite{zhang2025diffds} is a generative model capable of generating view-consistent 3D vector sketches either from a text description or a reference image, thus supporting both text-to-3D and image-to-3D generation. Essentially, it optimizes 3D rational Bezier curves using Score Distillation Sampling (SDS). \textbf{3Doodle}~\cite{10.1145/3658156} is an optimization-based method that generates descriptive and view-consistent sketch images given \emph{multi-view images} of a target object. It represents the sketch using 3D cubic Bezier curves (for view-independent lines) and superquadrics (for view-dependent contours). \textbf{Dream3DVG}~\cite{Li_2025_CVPR} is a text-to-vector graphics generation approach. It features a dual-branch framework that uses an auxiliary 3D Gaussian Splatting (3DGS~\cite{kerbl3Dgaussians}) branch to guide the 3D vector graphics optimization, enabling progressive coarse-to-fine detail refinement.

\keypoint{Evaluation Metrics.} We evaluate our method using three widely adopted metrics: semantic alignment, appearance alignment, and aesthetic quality. \textbf{Semantic Alignment (CLIP-S$_T$)}: To measure how well a generated 3D sketch matches the input text prompt $\mathcal{T}$, we adopt a multi-view CLIP-based similarity. Each 3D sketch is rendered from 16 fixed camera poses, and we compute the cosine similarity between the CLIP (\texttt{ViT-B/32})~\cite{radford2021clip} embedding of $\mathcal{T}$ and the embedding of each rendered view. The final score is the average similarity across all views. \textbf{Appearance Alignment (CLIP-S$_I$)}: When a reference image is provided, we assess how closely the generated sketch resembles it. We extract the CLIP image embedding of the reference and compute its average cosine similarity with the embeddings of the 16 rendered views. \textbf{Aesthetic Quality (AES)}: To evaluate visual appeal, we adopt a pre-trained aesthetic predictor~\cite{aesthetic_score} consisting of a frozen CLIP \texttt{ViT-L/14} encoder and an MLP head trained to regress human aesthetic judgments. Each rendered view is encoded into a 768-dimensional embedding, normalized, and fed to the MLP. The final aesthetic score is the mean prediction over the 16 views.

\subsection{Results}
\keypoint{Quantitative Results.} Table~\ref{table:main_results} compares our training-free 3DrawAgent (based on DeepSeek-V3.2 and Gemini-2.5Pro) against existing methods that require substantial model training. As shown, our approach delivers highly competitive performance across all metrics. In particular, 3DrawAgent achieves semantic alignment scores comparable to trained baselines for both category-level and fine-grained text descriptions. Additionally, our method consistently produces strong Aesthetic Scores, demonstrating the visual appeal and structural quality of the generated 3D sketches.
Importantly, all these results are obtained without any model fine-tuning, underscoring the effectiveness of our contrastive knowledge extraction pipeline in equipping frozen LLMs with robust 3D reasoning capabilities.

\keypoint{Qualitative Results.}
As shown in Figure~\ref{fig:qualitative} (a), our method (DeepSeek and Gemini) produces cleaner, more coherent, and more topologically accurate sketches than Diff3DS and Dream3DVG given category-level text prompts. Diff3DS often yields fragmented or chaotic curves, while Dream3DVG captures overall shapes but lacks structural precision. In contrast, our model consistently recovers canonical object geometry with clear part structure. Figure~\ref{fig:qualitative} (b) highlights our model's strength in handling complex, descriptive language. For prompts such as ``a fire phoenix engulfed in flames,'' baseline methods typically fail to represent key semantic elements, whereas our model leverages the LLM's compositional reasoning to jointly depict both the phoenix and associated flame structures, producing a coherent unified 3D form. In addition, as illustrated in Figure~\ref{fig:qualitative} (c), our image-to-3D results are significantly sparser and cleaner than Diff3DS. Our method, specifically 3DrawAgent (Gemini), accurately extracts major contours and structural lines from a single input image, yielding an interpretable and high-fidelity 3D wireframe.

\keypoint{More Results.} To complement our automated metrics, we conducted a user study evaluating Semantic Fidelity and Geometric Plausibility. Results demonstrate that our 3DrawAgent is significantly preferred, achieving a 46.66\% preference rate over Dream3DVG (36.67\%) and Diff3DS (16.67\%). User study details and failure cases are provided in the supplementary material.

\subsection{Ablation Study}
In this section, we conduct comprehensive ablation studies to validate the key components of our framework. Specifically, we examine: (i) the overall contribution of the experience library learned through our proposed Contrastive Knowledge Extraction (CKE), (ii) the effect of the contrastive group size $K$, and (iii) the necessity of ground-truth (GT) information. All results, reported in Table \ref{table:ablation}, use CLIP-S on the ModelNet40 test set.

\begin{table}[t]
\small
\begin{center}
\caption{Ablation study on the core components of our CKE pipeline. We report CLIP-S on the ModelNet40 test set. The Base model (Epoch 0) has no experience. Our method achieves strong results even without ground truth and benefits from a group size of $K=5$.}
\setlength{\tabcolsep}{2pt}
\label{table:ablation}
\scalebox{1.0}{
\begin{tabular}{l|l|ccccc}
\toprule
\bf Setting & \bf Component & \bf Ep 0 & \bf Ep 1 & \bf Ep 2 & \bf Ep 3 & \bf Ep 4 \\
\midrule
\rowcolor{lightgray}
\multicolumn{7}{l}{\textbf{1. Impact of Experience Library}} \\
Base & (w/o CKE) & 0.5735 & - & - & - & - \\
Ours & (w/ CKE) & 0.5735 & 0.6461 & \textbf{0.6643} & 0.6428 & 0.6416 \\
\midrule
\rowcolor{lightgray}
\multicolumn{7}{l}{\textbf{2. Impact of Group Size $K$}} \\
$K=2$ & (Test) & 0.5735 & 0.5947 & 0.6493 & 0.6466 & - \\
$K=5$ & (Test) & 0.5735 & 0.6461 & \textbf{0.6643} & 0.6428 & - \\
$K=10$ & (Test) & 0.5735 & 0.6135 & 0.5612 & 0.6148 & - \\
\midrule
\rowcolor{lightgray}
\multicolumn{7}{l}{\textbf{3. Impact of Ground Truth (GT)}} \\
GT=False & (Test) & 0.5735 & 0.6461 & \textbf{0.6643} & 0.6428 & 0.6416 \\
GT=True & (Test) & 0.5735 & \textbf{0.6648} & 0.6552 & 0.6141 & 0.6261 \\
\bottomrule
\end{tabular}
}
\vspace{-1cm}
\end{center}
\end{table}

\begin{figure*}
    \centering
    \includegraphics[width=\linewidth]{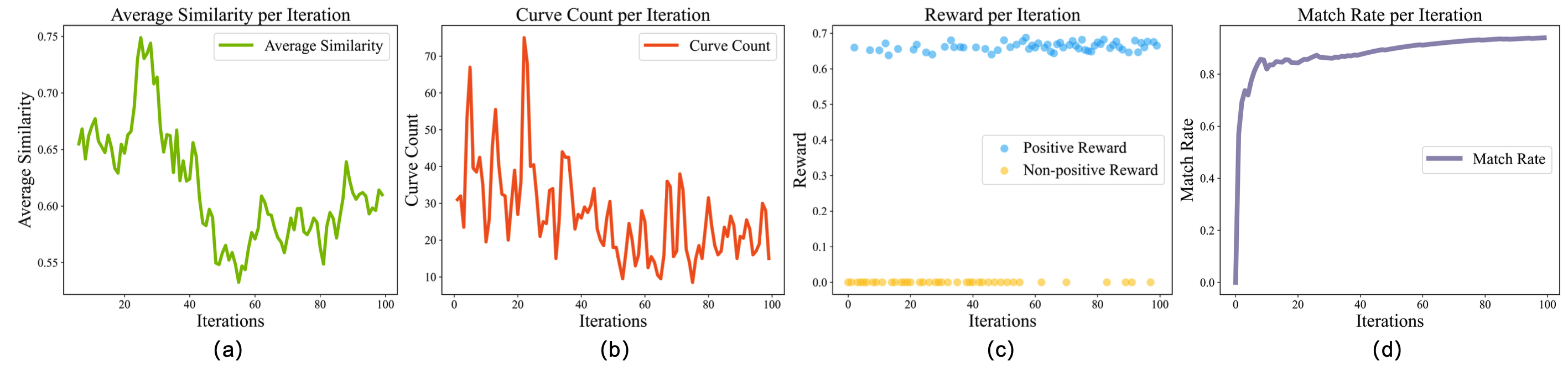}
    \caption{Statistics analysis of 200 rollouts for a single 3D drawing task during contrastive knowledge extraction, uniformly sampled to 100 for visualization. (a) Average pairwise similarity between curves within each rollout. (b) Distribution of curve counts across the 200 rollouts. (c) Reward score distribution over the 200 rollouts. (d) Bracket-matching rate, computed as matched cases divided by total cases.}
    \label{fig:train-stat}
\end{figure*}

\begin{figure*}[t]
    \centering
    \vspace{-0.5cm}
    \includegraphics[width=\linewidth]{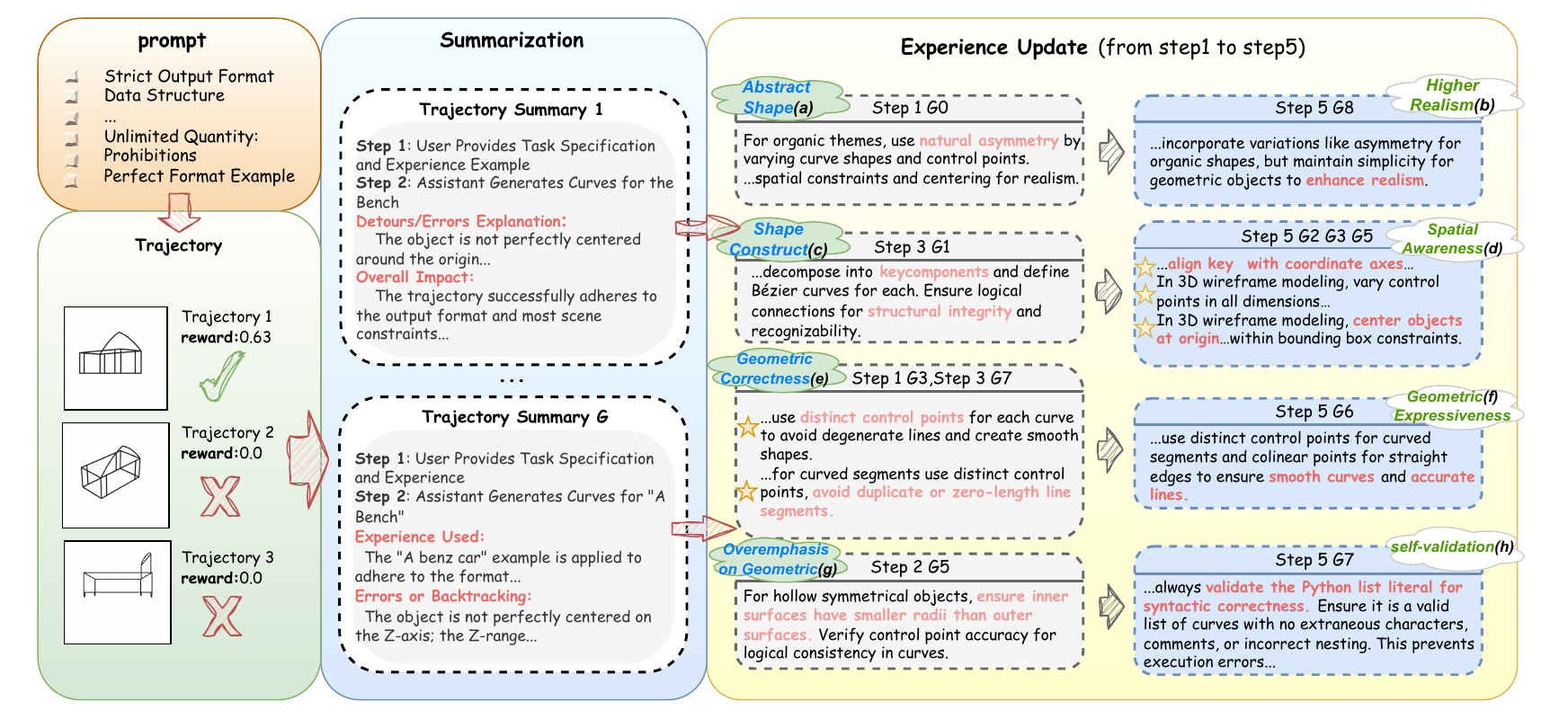}
    \vspace{-0.7cm}
    \caption{Extracted Knowledge Analysis.}
    \vspace{-0.6cm}
    \label{fig:semantic}
\end{figure*}

\keypoint{Impact of Learning Experience.}
We first compare our full 3DrawAgent model with a \emph{Base} model that uses the same LLM and in-context prompt but contains \emph{no learned experience}, i.e., w/o CKE. As shown in Table \ref{table:ablation} (Row 1), the Base model starts at a CLIP-S of 0.5735 (Epoch 0). After two CKE epochs, the score rises to 0.6643. This confirms that CKE effectively transforms noisy reward-based rollouts into concise, actionable 3D spatial principles, and that these accumulated experiences significantly enhance the LLM's generative quality. However, we can observe the performance initially rises and then slightly drops over epochs; we attribute this to the over-reasoning issue of LLMs.

\keypoint{Impact of Contrastive Group Size ($K$).}
Our framework relies on comparing $K$ candidate sketches for contrastive critique. We evaluate $K=2,5,10$. As shown in Table \ref{table:ablation} (Row 2), $K=5$ (our default) achieves the highest and most stable performance (0.6643). Increasing to $K=10$ brings no meaningful gain, while $K=2$ slows learning due to insufficient diversity. These results indicate that $K=5$ strikes an effective balance between informative contrast and computational efficiency.

\keypoint{Ground Truth Impact.}
We further examine whether CKE benefits from ground-truth supervision by comparing the setting without GT, i.e., \texttt{GT=False}, which relies solely on the multi-view CLIP reward, against the setting with GT provided during critique, i.e., \texttt{GT=True}. As shown in Table \ref{table:ablation} (Row 3), GT=True yields a faster early improvement (peaking at 0.6648 in Epoch 1), but GT=False reaches a nearly identical peak (0.6643) and learns more steadily over time. This demonstrates a key advantage of our framework: CKE remains highly effective even without ground-truth annotations, enabling scalable learning directly from reward signals alone.

\subsection{More Results and Analysis}
\keypoint{Statistics During Experience Extraction.} To understand the behavior of 3DrawAgent during contrastive knowledge extraction, we analyzed 200 rollout runs for a single task. As shown in Figure~\ref{fig:train-stat}, we find that low-reward outputs exhibit several patterns: (i) Curve degeneracy: many Bezier curves have nearly identical control points, resulting in straight, flat segments; (ii) Excessive curves: some sketches contain 500+ curves, adding redundancy without semantic gain; (iii) Polarized rewards: 30 outputs score 0.0, while 170 fall in [0.6, 0.7], providing clear contrastive signals; (iv) Structural irregularities: nested lists ([[[]]]) reduce readability and parsing robustness.

\keypoint{What's the Extracted Knowledge?} Here, we investigate how the LLM progressively improves its 3D sketch generation by analyzing the evolution of extracted experiences across multiple iterations of contrastive knowledge extraction (CKE). Figure~\ref{fig:semantic} reveals some interesting patterns: (i) \textbf{Semantic alignment} improves over time: the model initially generates abstract shapes (see Step1 G0 in Figure~\ref{fig:semantic}) but gradually learns to align orientations, thicknesses, and component relationships to better match the target theme, achieving \textbf{higher realism } (Step5 G8) and semantic consistency. (ii) Early-stage experiences focus primarily on \textbf{basic shape construction} (Step3 G1), such as component decomposition and symmetry. (iii) Over time, the focus shifts toward \textbf{spatial awareness} (Step5 G2), with experiences emphasizing full 3D control-point distributions, avoidance of planar collapse, and the representation of volumetric structures. (iv) Experiences around \textbf{control point usage} evolve from initially ensuring \textbf{geometric correctness} (Step3 G7), merely avoiding zero-length curves to enhance \textbf{geometric expressiveness} (Step5 G6) by specifying distinct points for curved segments and colinear points for straight edges, balancing smoothness and structural accuracy. (v) A key qualitative leap occurs when the model acquires \textbf{format self-validation} (Step5 G7) skills, such as verifying Python list syntax, nesting, and curve structure, ensuring outputs are both executable and robust. It compensates for the errors caused by an \textbf{overemphasis on geometric correctness} (Step2 G5).

\vspace{-0.3cm}
\section{Conclusion}
\vspace{-0.3cm}
We presented a training-free framework that enables LLMs to generate coherent 3D Bezier sketches through contrastive experience optimization. Unlike prior diffusion SDS-based methods that require explicit geometry supervision, our approach equips an off-the-shelf LLM with 3D spatial reasoning purely through self-produced rollouts and pairwise critique. Our experimental results show that learning drawing experiences are essential. Analysis of the extracted experiences further reveals a clear progression: from basic shape construction to full 3D spatial awareness, improved control-point usage, and robust output formatting. The results highlight a core insight: LLMs can acquire 3D geometric priors without parameter updates, simply by critiquing and refining their own outputs. We hope our work inspires broader training-free 3D reasoning and interactive tools using foundation models.

{
    \small
    \bibliographystyle{ieeenat_fullname}
    \bibliography{main}
}
\clearpage
\setcounter{page}{1}
\hypersetup{pageanchor=false}
\maketitlesupplementary
\appendix
\section*{Overview}

This supplementary material provides additional details, analyses, and results that further support the findings of our \textbf{3DrawAgent} framework. Specifically, it is organized as follows:

\begin{itemize}
    \setlength\itemsep{0.5em}

    \item \textbf{Section \ref{sec:implementation}} presents comprehensive \textbf{Implementation Details}, including renderer configurations, hyper-parameters for the adopted LLMs (DeepSeek and Gemini), and the settings of all evaluation metrics.

    \item \textbf{Section \ref{sec:more_results}} offers an extended analysis of \textbf{Stroke Count Constraints} and the \textbf{Variance in CKE}, illustrating how our method adapts to complexity budgets and demonstrating the necessity of CLIP-guided contrastive selection over random selection.

    \item \textbf{Section \ref{sec:user_study}} describes the design and outcomes of our \textbf{User Study}, which assesses the perceptual quality of our generated 3D sketches compared to baseline approaches.

    \item \textbf{Section \ref{sec:limitations}} provides a candid discussion on the \textbf{Limitations and Failure Cases} of our approach, analyzing common geometric challenges and outlining potential avenues for future research.

    \item \textbf{Section \ref{sec:prompts}} provides the detailed \textbf{Prompts} and execution logs used throughout our framework, followed by an extended gallery of 3D generation results across a wide range of object categories.

\end{itemize}

\section{Implementation Details}
\label{sec:implementation}

In this section, we present the detailed configurations used in our experiments, including the differentiable renderer setup, the CLIP-based evaluation metric, the Large Language Model (LLM) settings, and an analysis of the computational cost.

\subsection{Renderer and Evaluation Settings}
\label{subsec:renderer_settings}

\keypoint{Differentiable Renderer.}
Our rendering pipeline is built upon \texttt{pydiffvg}~\cite{li2020differentiable}. To ensure consistency across all experiments, we employ a unified \texttt{BatchRenderer} with the following fixed hyperparameters:
\begin{itemize}
    \item \textbf{Canvas \& Projection:} We render all 3D sketches onto a $512 \times 512$ canvas with perspective projection. The camera focal length is set to $907.32$ (derived from $\text{fov} \approx 60^{\circ}$ for a 512px width).
    \item \textbf{Curve Style:} The 3D Bezier curves are rasterized with a fixed stroke width of 2.0 pixels. The stroke color is set to dark gray (i.e., $\text{RGBA} = [0.1, 0.1, 0.1, 1.0]$) on a white background (i.e., $\text{RGBA} = [1.0, 1.0, 1.0, 1.0]$), composited via alpha blending.
    \item \textbf{Viewpoints:} We adopt a fixed set of \emph{16 camera poses} uniformly distributed around the object to capture multi-view geometric structure.
\end{itemize}

\keypoint{CLIP-based Scoring (CLIP-S).}

We utilize the pre-trained \texttt{ViT-B/32} model to measure text-3D sketch semantic alignment. Following Dream3DVG~\cite{Li_2025_CVPR}, for a given object category $C$ (e.g., ``car"), we construct a reference text prompt using a sketch-oriented template:

\begin{quote}
    \texttt{"\{$C$\}, minimal 2d line drawing, on a white background, black and white."}
\end{quote}

For each generated 3D sketch, we render 16 viewpoints and compute the cosine similarity between the embedding of text and each rendered view. The final CLIP-S score is the average similarity across all 16 views.

\subsection{LLM Configurations}
\label{subsec:llm_config}

We employ fixed hyperparameters for both Foundation Models (\texttt{DeepSeek-V3.2-Exp} and \texttt{Gemini-2.5 Pro}) to ensure a controlled balance between exploratory diversity during experience accumulation and deterministic behavior at inference.

\begin{itemize}
    \item \textbf{Exploration Phase (Training-Free CKE):} To promote diverse candidate sketches for contrastive critique, we set a sampling temperature of 0.7. The maximum output length is fixed at 32,768 tokens to support long chains of thought and large Bezier-curve lists. The GRPO contrastive group size is set to $K=5$.

    \item \textbf{Inference Phase:} For final 3D sketch generation, we reduce the sampling temperature to 0.3 for more stable and deterministic outputs, while keeping the token limit at 32,768 to preserve full-structure curve descriptions.

\end{itemize}

\begin{figure*}[t]
    \centering
    \includegraphics[width=\linewidth]{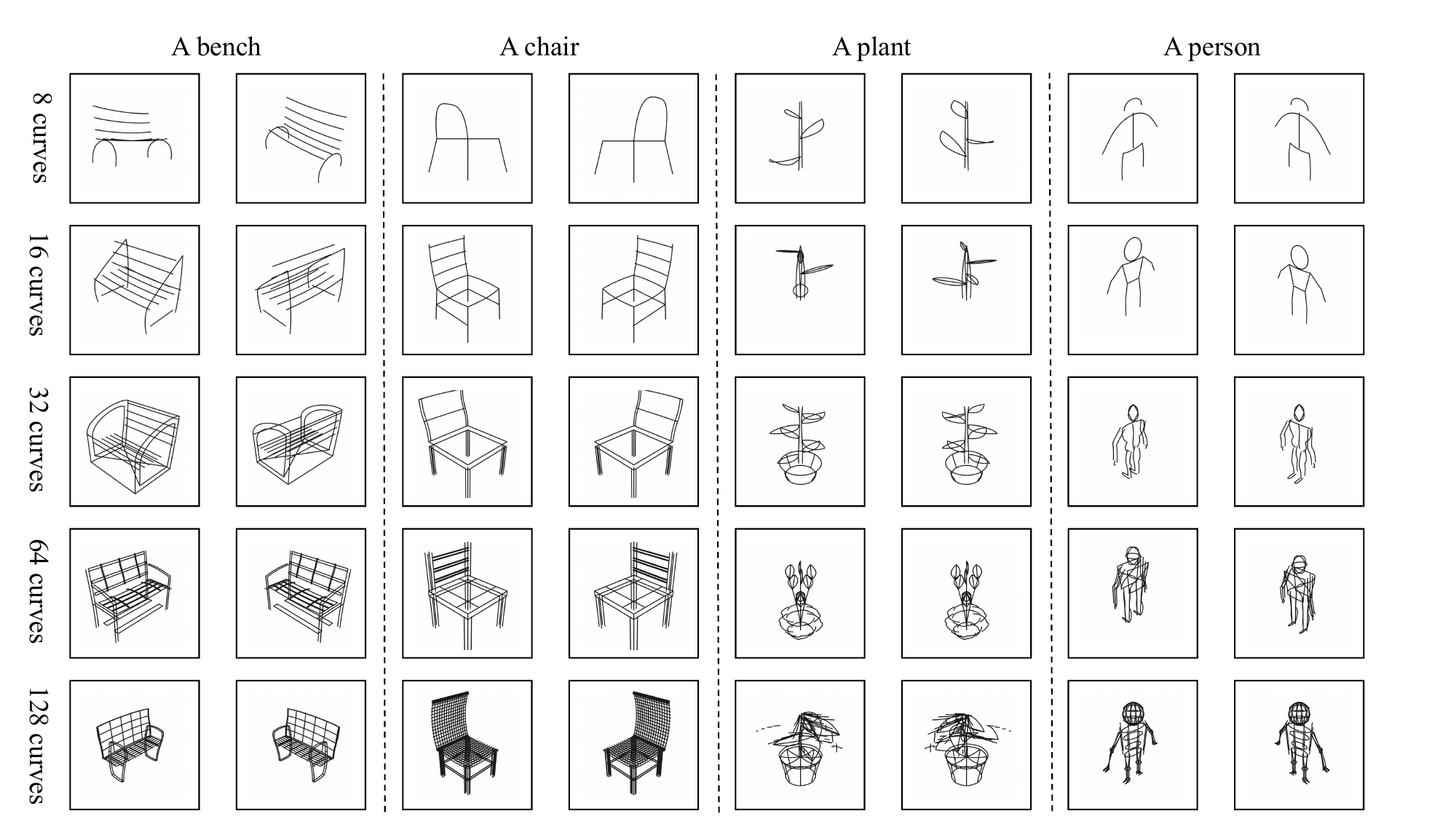}
    \caption{\textbf{Impact of Stroke Constraints on 3D Abstraction across Categories.}
    We evaluate the model's generation capability under varying Bezier curve budgets (rows from 8 to 128) across diverse categories: \texttt{Bench}, \texttt{Chair}, \texttt{Plant}, and \texttt{Person}.
    At minimal budgets (8 curves), the model performs high-level semantic abstraction, producing skeletal representations (e.g., a stick figure for the person or a simple stem for the plant).
    As the budget increases to 32--64 curves, structural details emerge, such as the pot geometry for the plant or parallel slats for the furniture.
    At 128 curves, the sketches evolve into dense wireframes.
    This demonstrates the model's versatility in adapting its planning strategy from abstract symbolism to geometric fidelity for both rigid and organic shapes.}
    \label{fig:stroke_analysis}
\end{figure*}

\subsection{Computational Cost and Efficiency}
\label{subsec:cost_analysis}

Unlike optimization-based methods (e.g., SDS) that require per-instance gradient updates, our framework is training-free and relies on API-based LLM inference. Below, we report the empirical computational cost measured using \texttt{DeepSeek-V3.2-Exp}.

\keypoint{Cost Comparison.}
Table \ref{tab:cost_comp} compares the inference latency and average monetary cost per sample of our method against state-of-the-art optimization-based baselines. While prior methods require 60 to 120 minutes of expensive GPU computation per object, our training-free, API-based approach drastically reduces the generation time to approximately 2 minutes per sample, lowering the average cost to just \$0.09.

\begin{table}[t]
    \centering
    \caption{Cost comparison with single objects.}
    \label{tab:cost_comp}
    \resizebox{\linewidth}{!}{
    \begin{tabular}{l|ccc}
        \toprule
        \bf Method & \bf GPU / API & \bf min / Sample & \bf Avg. Cost (USD)\\
        \hline
        \bf 3Doodle & NVIDIA RTX3090 & $\sim$ 120 & 0.80 \\
        \bf Diff3DS & NVIDIA A10 & $\sim$ 120 & 1.50 \\
        \bf Dream3DVG & NVIDIA A100 & $\sim$ 60 & 1.30 \\
        \bf Ours & DeepSeek-V3.2 & $\sim$ 2 & 0.09 \\
        \bottomrule
    \end{tabular}
    }
\end{table}

\keypoint{Monetary Cost for CKE.}
To further detail our API consumption, we conducted a full CKE (Contrastive Knowledge Extraction) run on a dataset of 100 prompts over 3 epochs with a group size of $K=5$. The total API cost was $\approx$ \$11 USD. The detailed pricing model and estimated consumption are listed in Table \ref{tab:api_cost}.

\begin{table}[t]
    \centering
    \caption{\textbf{Training-Free Cost Analysis (DeepSeek-V3.2-Exp).} Costs are estimated for a complete experience extraction run (100 prompts, 3 epochs, $K=5$).}
    \label{tab:api_cost}
    \resizebox{\linewidth}{!}{
    \begin{tabular}{lccc}
        \toprule
        \textbf{Metric} & \textbf{Unit Price (USD)} & \textbf{Total Volume} & \textbf{Est. Cost (USD)} \\
        \midrule
        Input (Cache Hit) & 0.027 / 1M tokens & $\sim$150M tokens & 4.1 \\
        Input (Cache Miss) & 0.275 / 1M tokens & $\sim$10M tokens & 2.8 \\
        Output & 0.413 / 1M tokens & $\sim$10M tokens & 4.1 \\
        \midrule
        \textbf{Total} & - & - & \textbf{11.0} \\
        \bottomrule
    \end{tabular}
    }
\end{table}

\section{More Results and Analysis}
\label{sec:more_results}

In this section, we provide further analysis of the model's behavior and learning stability. Specifically, we evaluate its controllability over abstraction levels via stroke count constraints, and investigate the variance of our Contrastive Knowledge Extraction (CKE) compared to a random selection baseline.

\subsection{Abstract Level with Stroke Number Control}
\label{subsec:stroke_count}

A key advantage of our language-driven framework, relative to pixel-space or fixed-representation generative methods, is its explicit controllability over the abstraction level via natural-language constraints. By specifying the desired number of curves (e.g., ``draw a bench using exactly 16 curves"), the LLM is encouraged to allocate its limited geometric budget toward semantically important structures.

To evaluate this controllability, we prompt the model to sketch a \texttt{bench} under different stroke-count constraints: 8, 16, 32, 64, and 128 curves. The resulting sketches are visualized in Figure~\ref{fig:stroke_analysis}.

\keypoint{Abstraction vs. Detail.}
As shown in Figure \ref{fig:stroke_analysis}, under a tight budget of 8 curves (i.e., the first row), the model demonstrates emergent reasoning by focusing on the most essential components. Curves are primarily allocated to outline the seat and the four legs, while finer details and textures are omitted. This behavior indicates that the LLM encodes an internal hierarchy of shape semantics, prioritizing structural integrity over decorative elements.

\keypoint{Progressive Refinement.} With a higher curve budget of 16 and 32, the model gradually transitions from a \emph{skeletal} to a more \emph{descriptive} representation. Additional curves are allocated to the backrest and seat, capturing details such as the slats of a wooden bench. This smooth and coherent refinement demonstrates that the learned experience library $\mathcal{E}$ effectively guides the model in managing increased complexity while maintaining geometric consistency.

\keypoint{High-Density Generation.}
With 64 and 128 curves (i.e., the last two rows in Figure \ref{fig:stroke_analysis}), the sketches form dense wireframes. Unlike standard mesh reconstruction methods that can struggle with topology, our approach preserves clean vector curves. However, beyond a certain point (e.g., 128 curves), perceptual improvement plateaus, and the model may introduce redundant or overlapping lines to exhaust the curve budget. This demonstrates the trade-off between efficiency and fidelity, suggesting that a medium budget (32--64 curves) typically provides the optimal balance for concept design tasks.

\subsection{Variance in CKE and Random Selection}
\label{subsec:random_selection}

To further validate the effectiveness of our CLIP-guided Contrastive Knowledge Extraction (CKE), we compare it against a random selection baseline. In the random selection setting, instead of forming contrastive pairs based on multi-view CLIP similarity scores, we randomly sample generated sketches to form ``relatively better'' and ``worse'' pairs for the LLM to critique. The comparison of semantic alignment (CLIP-S$_T$) over multiple epochs is presented in Table \ref{tab:random_selection}.

\begin{table}[t]
    \centering
    \caption{\textbf{Comparison of CKE against Random Selection.} We report the CLIP-S$_T$ scores across different experience extraction epochs.}
    \label{tab:random_selection}
    \resizebox{\linewidth}{!}{
    \begin{tabular}{llcccc}
        \toprule
        \textbf{Setting} & \textbf{Component} & \textbf{Ep 0} & \textbf{Ep 1} & \textbf{Ep 2} & \textbf{Ep 3} \\
        \midrule
        Base & (w/o CKE) & 0.5735 & - & - & - \\
        Random & (w/ CKE) & 0.5735 & 0.6420 & 0.5595 & 0.6094 \\
        \textbf{Ours} & \textbf{(w/ CKE)} & 0.5735 & 0.6461 & \textbf{0.6643} & 0.6428 \\
        \bottomrule
    \end{tabular}
    }
\end{table}

\keypoint{Impact of Random Pairs.}
As shown in Table \ref{tab:random_selection}, random pair selection leads to highly unstable performance across CKE iterations, with a significant performance drop in Epoch 2 (0.5595, which is even lower than the Base model's 0.5735). We attribute this instability to the fact that randomly sampled pairs often lack a clear semantic ordering, producing noisy and sometimes contradictory preference signals. In contrast, our CLIP-guided selection provides reliable and consistent guidance, allowing the model to steadily accumulate beneficial spatial principles and achieve consistent gains.

\keypoint{CKE Variance and Over-reasoning.}
Table \ref{tab:random_selection} also reveals a slight performance drop for our method in Epoch 3 (from 0.6643 down to 0.6428). As CKE progressively extracts and integrates experiences over multiple iterations, the newly extracted rules in later stages can sometimes become local, task-biased, or overly specific. When these overly specific constraints are integrated into the experience bank, they may reduce the LLM's drawing flexibility or cause ``over-reasoning,'' which slightly degrades its generalization to novel prompts. This observation highlights the importance of maintaining a concise, abstract, and high-level experience library.

\section{User Study}
\label{sec:user_study}

To better evaluate the perceptual quality of the generated 3D sketches, we conducted a user study comparing our \textbf{3DrawAgent} against two state-of-the-art baselines: \textbf{Diff3DS}~\cite{zhang2025diffds} and \textbf{Dream3DVG}~\cite{Li_2025_CVPR}.

\keypoint{Participants and Dataset.}
We recruited 30 volunteers, primarily university students and researchers with backgrounds in computer science and design (ages 18--28), with a gender ratio of roughly 5:1 (male to female). The evaluation set comprised 40 randomly selected prompts spanning both rigid objects (e.g., furniture, vehicles) and organic shapes (e.g., animals, plants), consistent with the categories presented in the qualitative comparisons of the main paper.

\keypoint{Procedure and Criteria.}
For each prompt, participants were shown three anonymized 3D sketches, i.e., rendered as rotating videos to display full 360-degree structure, generated by Diff3DS, Dream3DVG, and our method. The presentation order was randomized to avoid bias. Participants were asked to select the best sketch based on two criteria:
\begin{itemize}
    \item \textbf{Semantic Fidelity:} How accurately the sketch reflects the input text description.
    \item \textbf{Geometric Plausibility:} Whether the 3D structure is coherent, clean, and free of floating artifacts or fragmented curves.
\end{itemize}

\begin{figure}
    \centering
    \includegraphics[width=\linewidth]{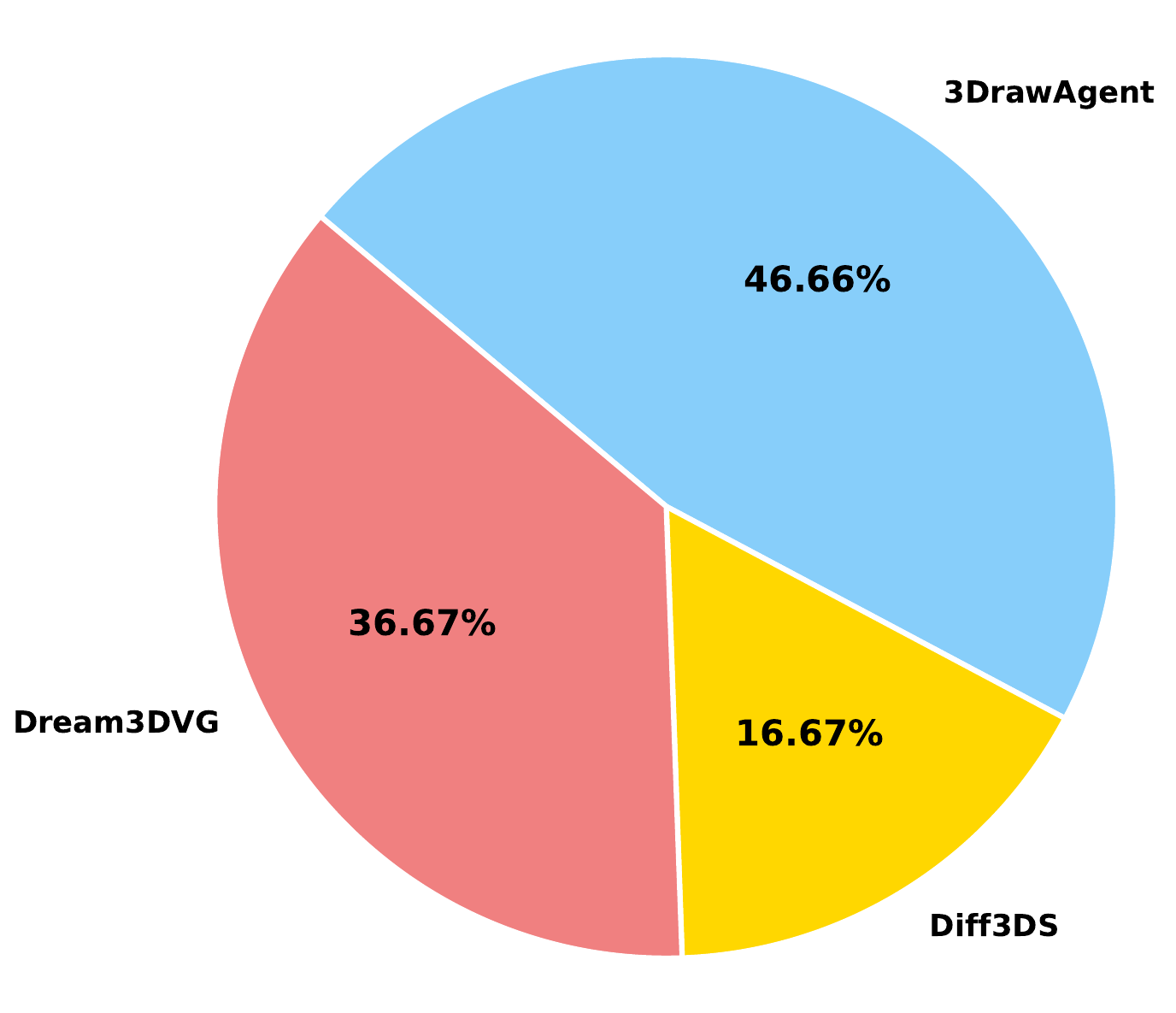}
    \caption{\textbf{User Study Results.} Percentage of user preference votes. \textbf{3DrawAgent} (46.66\%) is the most preferred method, showing a clear advantage over \textbf{Dream3DVG} (36.67\%) and \textbf{Diff3DS} (16.67\%) in terms of combined semantic and geometric quality.}
    \label{fig:user_study_pie}
\end{figure}

\keypoint{Results.}
The results of the user study are summarized in Figure \ref{fig:user_study_pie}. Our method, \textbf{3DrawAgent}, achieved the highest preference rate with \textbf{46.66\%} of votes. \textbf{Dream3DVG} followed with \textbf{36.67\%}, while \textbf{Diff3DS} received \textbf{16.67\%}. Participants noted that \textbf{Dream3DVG} often captures overall object volume well, benefiting from its 3DGS guidance, but occasionally produces over-smoothed or noisy strokes. In contrast, \textbf{3DrawAgent} was consistently praised for its clean, \emph{designer-like} vector curves and superior structural logic, particularly in handling complex topologies where explicit geometric reasoning is critical.

\section{Limitations and Failure Cases}
\label{sec:limitations}

\begin{figure}[t]
    \centering
    \includegraphics[width=\linewidth]{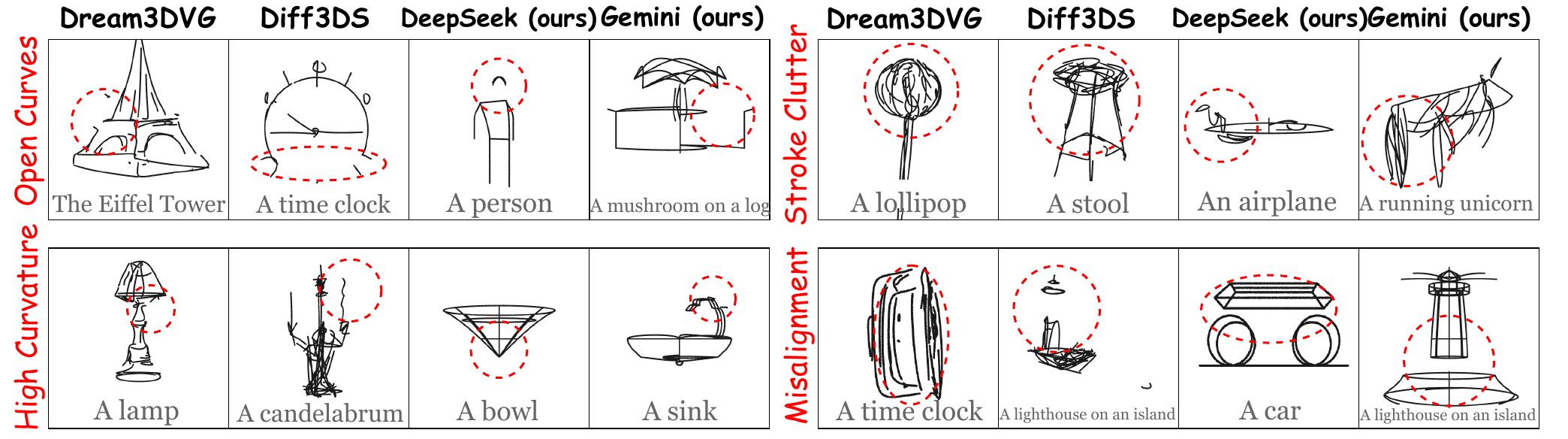}
    \caption{\textbf{Visual Examples of Common Failure Modes.} Despite our experience bank's guidance, 3DrawAgent encounters challenges with strict geometric connectivity and handling semantic ambiguity in complex structures. Key issues include (a) disconnected junctions where strokes should intersect, (b) floating components, and (c) visual clutter when managing ambiguous topological constraints.}
    \label{fig:failures}
\end{figure}

Despite 3DrawAgent's capacity to generate semantically accurate and abstract 3D sketches training-free, our method exhibits certain limitations common to parametric curve generation via text-guided optimization. Below, we dissect specific failure modes and propose directions for future refinement, referencing visual examples in Figure~\ref{fig:failures}.

\keypoint{Strict Geometric Plausibility and Connectivity.}
A primary challenge lies in enforcing strict geometric constraints, such as perfect connectivity between adjacent curves that form semantic joints (e.g., where table legs meet the table top). While the extracted experience bank $\mathcal{E}$ provides high-level structural guidance (e.g., ``legs should be placed vertically under the table top"), it lacks a dense vector point supervision to enforce localized endpoint intersection. As visualized in Figure~\ref{fig:failures} (a) and (b), strokes corresponding to different semantic components may fail to intersect precisely, resulting in slightly disconnected joints or ``floating" elements. The current loss function, primarily a holistic CLIP-based similarity score, optimizes for overall semantic recognition rather than local topological exactness. Incorporating explicit intersection-promoting or endpoint-matching loss terms during optimization could mitigate this issue.

\keypoint{Semantic Ambiguity and Component Placement.}
As a text-driven agent relying on a frozen, generalized geometric model, 3DrawAgent sometimes struggles with ambiguous semantic placement of geometric components, particularly for non-canonical object structures. As shown in Figure~\ref{fig:failures} (a), while the semantics of a ``stool" are captured, the spatial relationships between the individual curves defining the support legs are geometrically loose. In Figure~\ref{fig:failures} (c), when prompted to sketch a ``stroller," the agent generates a plausible overall structure but creates complex, overlapping line-clusters rather than clean contours for detailed components (like wheels), leading to visual clutter.

\keypoint{Future Work Direction.}
These failure cases highlight that high-level linguistic reasoning alone is insufficient for precise geometric reasoning. Future research could focus on two avenues: (1) integrating learned geometric priors (e.g., pre-trained wireframe reconstruction models) into the generation pipeline to impose better local structural order, or (2) developing dense, multi-view reward functions that specifically penalize floating primitives or incomplete geometric loops.

\section{Prompts and More Results}
\label{sec:prompts}

In this section, we provide the exact prompt specifications and execution logs used in our framework \textbf{3DrawAgent}. To ensure full reproducibility, we present the raw content of our System Prompt with illustrative generation logs that highlight the step-by-step execution and output behavior. Finally, we present an extended gallery of qualitative results to demonstrate the robust zero-shot generalization of our method.

\subsection{System Prompt Specification}
\label{subsec:system_prompt}

\begin{figure*}[t]
    \centering
    \includegraphics[width=\linewidth]{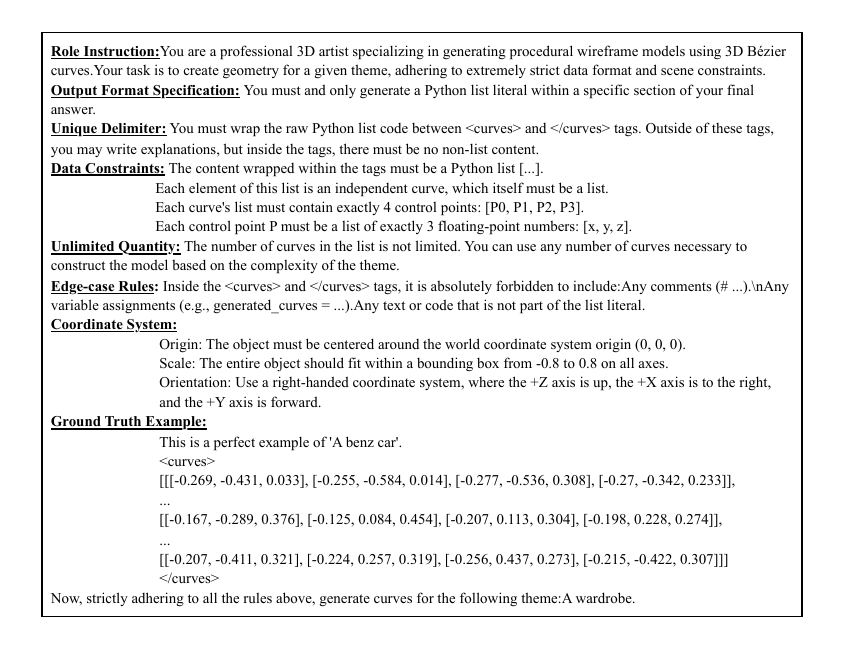}
    \caption{\textbf{Full System Prompt.} Raw text input provided to the LLM, combining role definition, strict syntax constraints (code-only output), coordinate system rules, and a few-shot example (``A benz car") to guide 3D sketch generation.}
    \label{fig:full_system_prompt}
\end{figure*}

Figure \ref{fig:full_system_prompt} shows the \emph{Input to the Agent (LLM)}. This prompt is a comprehensive instruction set designed to initialize the LLM as a 3D spatial planner. It serves several critical roles in our framework:

\begin{itemize}
    \item \textbf{Role \& Format Definition:} The ``Role Instruction" and ``Output Format Specification" constrain the LLM's output to a Python list of 3D coordinates, ensuring deterministic parsing by the renderer without syntax errors.
    \item \textbf{Coordinate Grounding:} The ``Coordinate System"
    defines physical bounds ($[-0.8, 0.8]$) and orientation (Right-handed, Z-up), providing a geometric prior that prevents out-of-view or distorted generations.
    \item \textbf{In-Context Learning:} The ``Ground Truth Example" (e.g., \texttt{a Benz car}) provides a dense, high-quality reference. This allows the LLM to internalize the expected \textit{density} and \textit{topology} of curves before generating new targets (e.g., \texttt{A wardrobe}).
\end{itemize}

\subsection{Generation Logs}
\label{subsec:gen_logs}

\begin{figure*}[t]
    \centering
    \includegraphics[width=\linewidth, height=0.92\textheight, keepaspectratio]{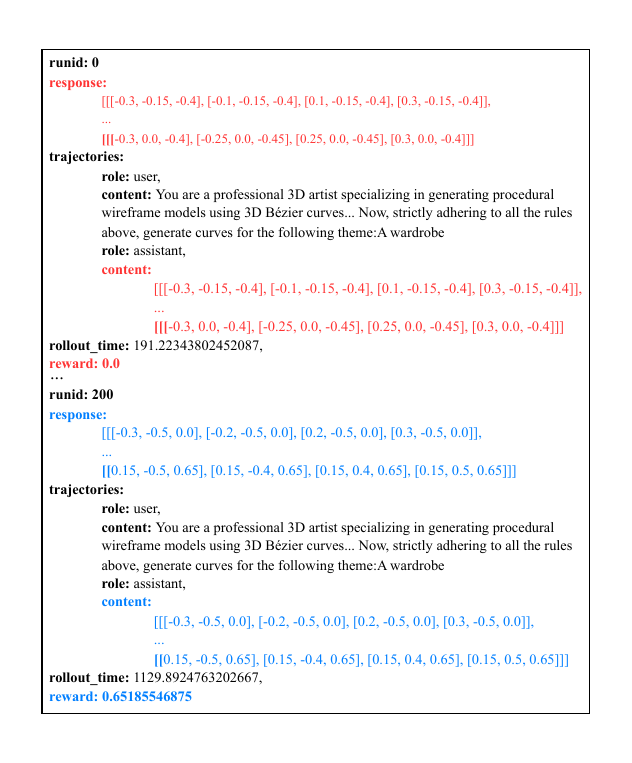}
    \caption{\textbf{Data Flow in Experience Extraction.} Raw logs collected during exploration, capturing the Agent's outputs (3D curves) and the Environment's feedback (rewards). These paired samples serve as the \emph{input} to the ``Contrastive Knowledge Extraction" module, where the ``Judge LLM" derives spatial principles by contrasting high- and low-reward trajectories.}
    \label{fig:run_logs}
\end{figure*}

    Figure \ref{fig:run_logs} shows the raw logs, which serve as a key data carrier in our framework, during 3D drawing using LLM. Beyond indicating success or failure, these logs capture the intermediate states of our training-free loop, functioning both as the \emph{output of the exploration phase} and the \emph{input to the reflection phase}.

\begin{itemize}
    \item \textbf{Source (Agent-Environment Interaction):} The logs capture the LLM's interaction with the evaluator, i.e., the recorded trajectory containing the user prompt and the assistant's response. The \texttt{response} field stores the 3D curve coordinates generated by the Agent, while the \texttt{reward} field provides the feedback computed by the Environment (CLIP-based renderer).

    \item \textbf{Destination (Input to Judge):} These logs are fed into the Judge LLM (i.e., Experience/Knowledge Extractor), which contrasts high-scoring entries (e.g., \texttt{runid: 200}, Reward 0.65) with low-scoring ones (e.g., \texttt{runid: 0}, Reward 0.0) to perform causal reasoning.

    \item \textbf{Algorithmic Function:} The contrastive analysis allows the system to self-diagnose: failures in low-reward runs are attributed to \emph{geometric sparsity} (i.e., short rollouts, simple curve lists), whereas successes arise from \emph{dense spatial planning} (i.e., longer inference, utilization of Z-axis). These insights are distilled into textual \emph{Experiences} to guide future 3D sketch generations.
\end{itemize}

\subsection{Additional Qualitative Results}
\label{subsec:more_qualitative}

\begin{figure*}[t]
    \centering
    \includegraphics[width=\linewidth, height=0.95\textheight, keepaspectratio]{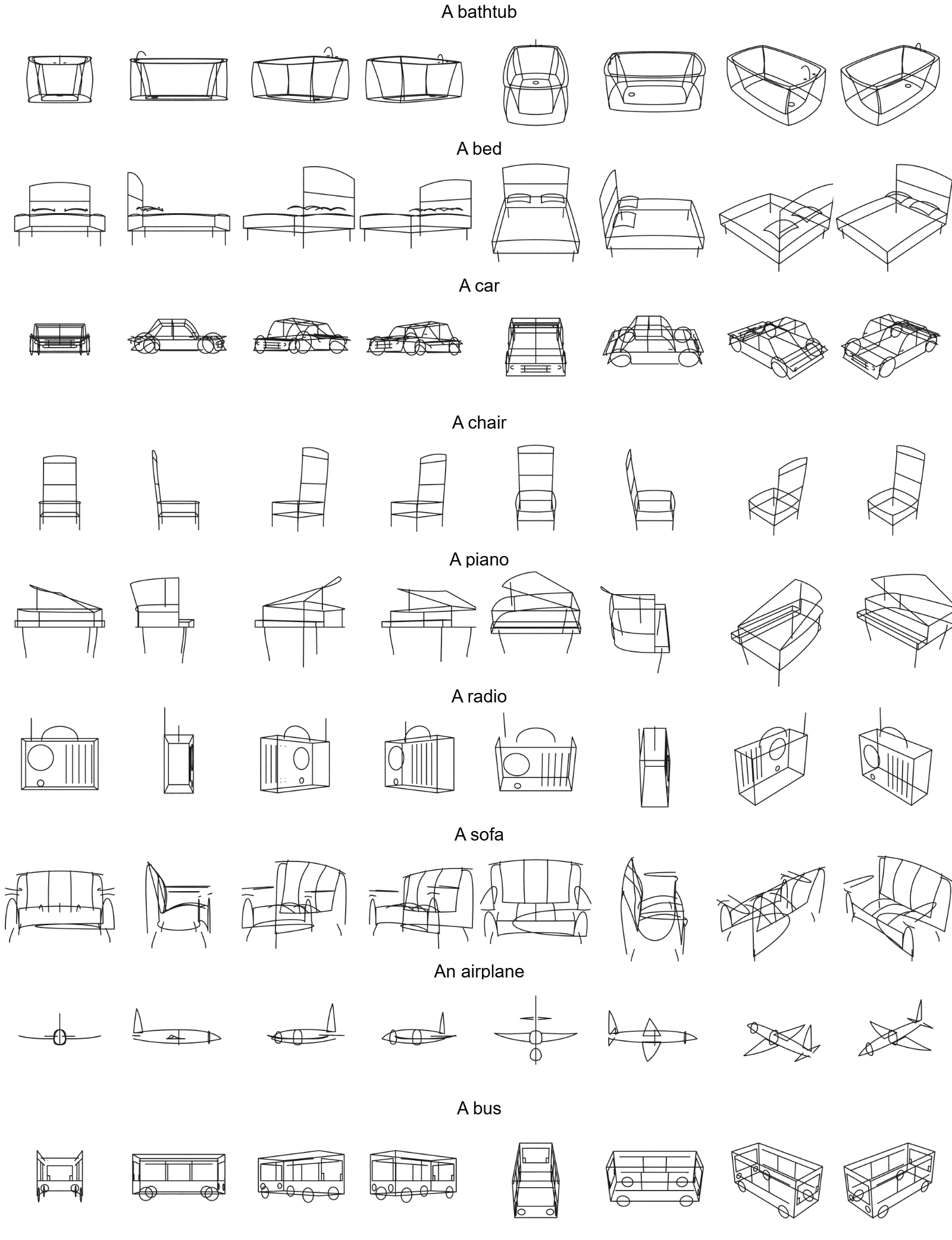}
    \caption{Additional Results: Text-to-3D Generation (Simple Prompts).}
    \label{fig:add_text_simple}
\end{figure*}

\begin{figure*}[t]
    \centering
    \includegraphics[width=\linewidth, height=0.95\textheight, keepaspectratio]{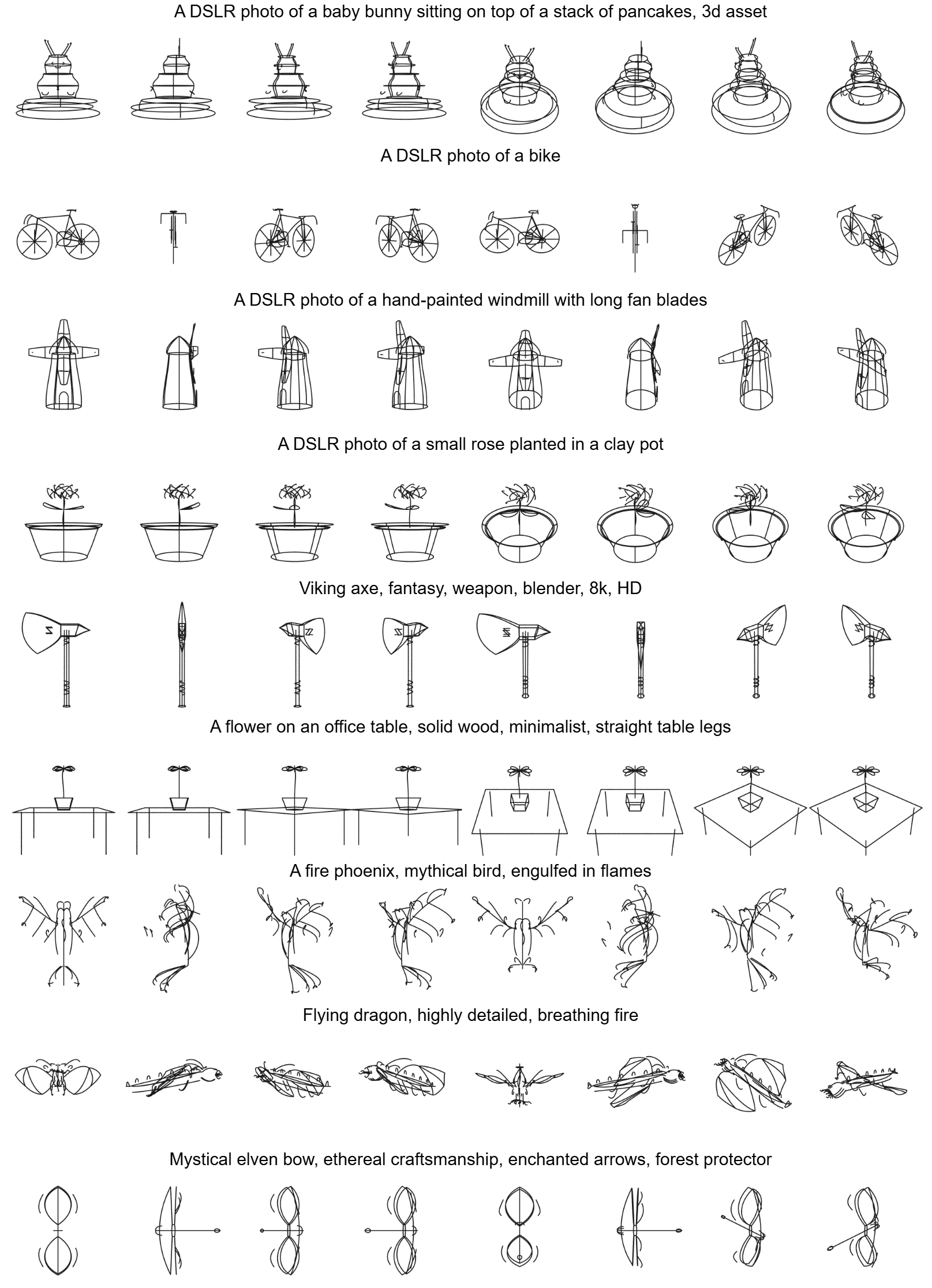}
    \caption{Additional Results: Text-to-3D Generation (Complex Prompts).}
    \label{fig:add_text_complex}
\end{figure*}

\begin{figure*}[t]
    \centering
    \includegraphics[width=\linewidth, height=0.95\textheight, keepaspectratio]{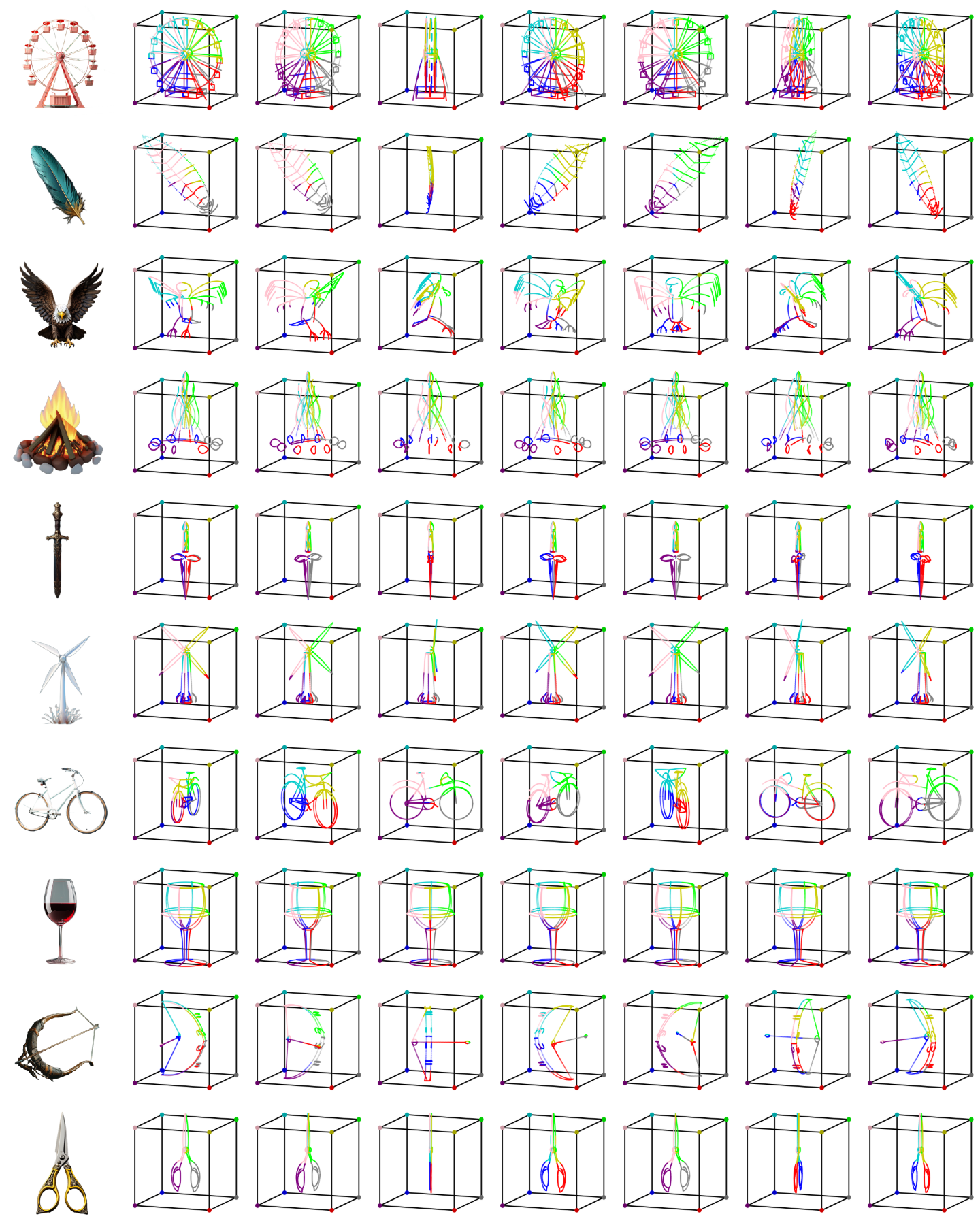}
    \caption{Additional Results: Image-to-3D Generation (Part I).}
    \label{fig:add_image_part1}
\end{figure*}

\begin{figure*}[t]
    \centering
    \includegraphics[width=\linewidth, height=0.95\textheight, keepaspectratio]{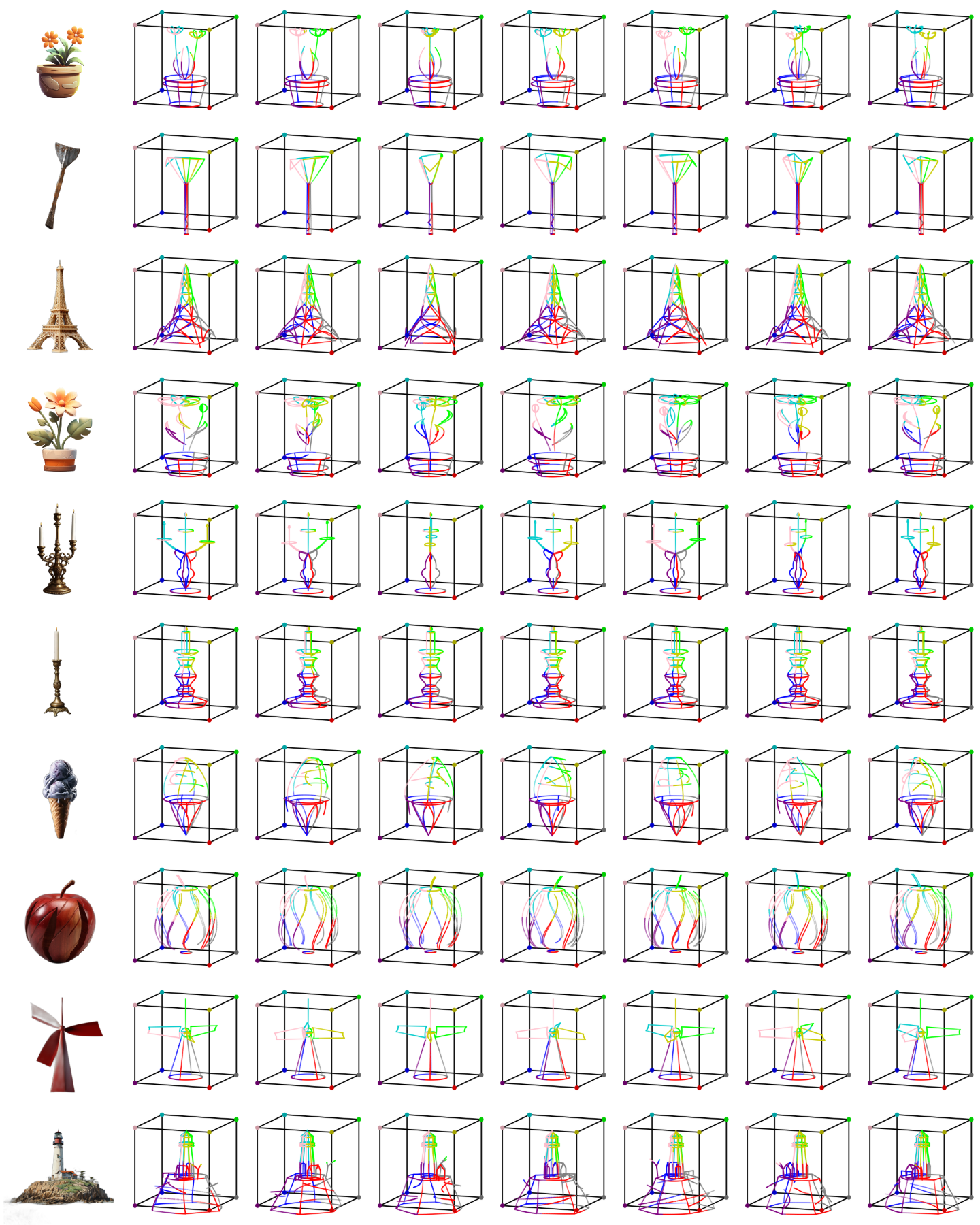}
    \caption{Additional Results: Image-to-3D Generation (Part II).}
    \label{fig:add_image_part2}
\end{figure*}

In addition to the analysis above, we present more 3D generation results across diverse categories to demonstrate the robustness of our method, as shown in Figures~\ref{fig:add_text_simple}, \ref{fig:add_text_complex}, \ref{fig:add_image_part1}, and \ref{fig:add_image_part2}.

\end{document}